\documentclass[runningheads]{llncs}

\usepackage{eccv}

\usepackage{eccvabbrv}

\usepackage{graphicx}
\usepackage{booktabs}

\usepackage[accsupp]{axessibility}  

\usepackage{hyperref}

\usepackage{orcidlink}

\usepackage{ragged2e} 
\usepackage{utfsym}
\usepackage{makecell, multirow, tabularx}
\usepackage{bm}
\usepackage{amssymb,amsmath}
\usepackage{enumitem}
\usepackage{bbm}
\usepackage{bbding}
\usepackage{tipa}
\usepackage{stfloats}
\usepackage{floatrow}
\newfloatcommand{capbtabbox}{table}[][\FBwidth]
\usepackage{tabularx}

\hypersetup{
 colorlinks=true,
 linkcolor=red,
 citecolor=cyan,
 filecolor=magenta, 
 urlcolor=cyan,
 }
\begin{document}

\title{KMTalk: Speech-Driven 3D Facial Animation with Key Motion Embedding} 

\titlerunning{KMTalk}

\renewcommand{\thefootnote}{\fnsymbol{footnote}}

\author{Zhihao Xu\inst{1\footnotemark[1]}\and
Shengjie Gong\inst{1\footnotemark[1]}\and
Jiapeng Tang\inst{2}\and
Lingyu Liang\inst{1}\and
Yining Huang\inst{1}\and
Haojie Li\inst{1}\and
Shuangping Huang\inst{1,3\footnotemark[2]}}

\authorrunning{Z.~Xu et al.}

\institute{South China University of Technology\\
\and
Technical University of Munich\\
\and
Pazhou Laboratory\\
\email{\{eezhihaoxu,eeshengjiegong\}@mail.scut.edu.cn},
\email{eehsp@scut.edu.cn}
}

\maketitle
\footnotetext[1]{Authors contributed equally.}
\footnotetext[2]{Corresponding author.}
\begin{abstract}
We present a novel approach for synthesizing 3D facial motions from audio sequences using key motion embeddings. Despite recent advancements in data-driven techniques, accurately mapping between audio signals and 3D facial meshes remains challenging. Direct regression of the entire sequence often leads to over-smoothed results due to the ill-posed nature of the problem. To this end, we propose a progressive learning mechanism that generates 3D facial animations by introducing key motion capture to decrease cross-modal mapping uncertainty and learning complexity.  Concretely, our method integrates linguistic and data-driven priors through two modules: the linguistic-based key motion acquisition and the cross-modal motion completion. The former identifies key motions and learns the associated 3D facial expressions, ensuring accurate lip-speech synchronization. The latter extends key motions into a full sequence of 3D talking faces guided by audio features, improving temporal coherence and audio-visual consistency. Extensive experimental comparisons against existing state-of-the-art methods demonstrate the superiority of our approach in generating more vivid and consistent talking face animations. Consistent enhancements in results through the integration of our proposed learning scheme with existing methods underscore the efficacy of our approach. Our code and weights will be at the project website: \textcolor{magenta}{\url{https://github.com/ffxzh/KMTalk}}.
\end{abstract}
\section{Introduction}
\label{sec:intro}
Speech-driven 3D facial animation aims to create realistic talking heads that synchronize with input speech.  It plays a significant role in many applications of virtual reality, like film production, computer gaming, and education \cite{tanaka2022acceptability,liu2009analysis}.

\begin{figure}[t]
    \centering
    \includegraphics[width=\linewidth]{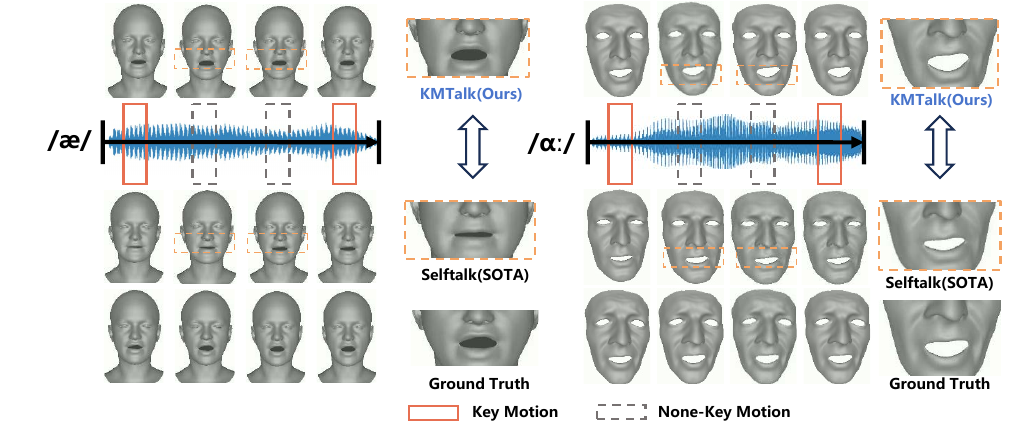}
    \caption{Compared to the state-of-the-art method Selftalk, our approach can produce more vivid lip motions from speeches, since we introduce linguistic priors to characterize key motions and utilize data-driven priors to interpolate non-key motions.
    }
    \label{fig:fig1}
\end{figure}

The main challenge of speech-driven 3D talking faces lies in the ill-posed problem caused by the cross-modal mapping uncertainty from the speech domain to the 3D motion domain. Since there may be multiple plausible outputs for input audio, effective regularizations and constraints should be integrated into the system, to generate vivid facial motions. The related methods can be roughly divided into linguistic-based methods~\cite{edwards2016jali,massaro201212,taylor2012dynamic,xu2013practical,bao2023learning} and data-driven methods~\cite{cudeiro2019capture,fan2022faceformer,habibie2021learning,liu2015video, wang20213d,xing2023codetalker,peng2023selftalk, richard2021meshtalk, wu2023speech, nocentini2023learning}.
For linguistic-based methods~\cite{edwards2016jali,massaro201212,taylor2012dynamic,xu2013practical,bao2023learning}, a set of intricate phoneme-to-viseme mapping rules is manually designed to generate the talking mouth based on priors from visemes or linguistic knowledge.
%
While these methods explicitly control the animation of articulation processes, such as procedural lip sync with animation curves, 
their focus is mainly on localized facial movements, like those of the mouth area, lacking a systematic approach for modeling comprehensive facial motion.
Thanks to the established audio-to-face datasets, learning-based methods ~\cite{fan2022faceformer,habibie2021learning,wang20213d,xing2023codetalker,peng2023selftalk,richard2021meshtalk,wu2023speech,nocentini2023learning} choose to map audio signals into 3D facial meshes in a data-driven manner. Most of these works~\cite{cudeiro2019capture,fan2022faceformer,habibie2021learning,liu2015video, wang20213d,xing2023codetalker,peng2023selftalk, richard2021meshtalk,wu2023speech,nocentini2023learning,wang2023seeing} typically formulated the cross-modal mapping of 3D talking face generation as a regression task,
such as MeshTalk~\cite{richard2021meshtalk}, FaceFormer~\cite{fan2022faceformer}, and SelfTalk~\cite{peng2023selftalk}.
While achieving impressive performance, they exhibit common limitations in their learning schemes.
Firstly, they directly learn the ambiguous cross-modal mapping between audio and facial expression sequences, always leading to sub-optimal results in terms of temporal coherence and audio-visual consistency. 
Secondly, these methods typically regress the entire sequence without considering key motion cues, hindering the capture of detailed facial dynamics and accurate lip movements, particularly in complex facial expressions such as puckering or opening the mouth (as depicted in Fig.~\ref{fig:fig1}). 
Lastly, they overlook linguistic priors essential for simulating the articulation process, thereby limiting their ability to achieve precise lip-speech synchronization.

To this end, inspired by the keyframe-based video generation techniques observed in recent studies~\cite{yin2023nuwa,lu2022video,wen2018generating,niklaus2017video}, which prioritize the generation of keyframes before adding detailed elements, 
we introduce a progressive learning mechanism that generates realistic 3D facial animations from audio inputs by incorporating key motion embeddings. 
The key idea is to initially generate key facial expressions, and then interpolate the intermediate motions to obtain the entire motion sequence, which significantly reduces the uncertainty of cross-modal mapping and eases the learning difficulty. 
Concretely, our method integrates linguistic and data-driven priors through two modules: the linguistic-based key motion acquisition and the cross-modal motion completion.
The linguistic-based key motion acquisition module utilizes phoneme-based localization methods to identify temporal indices of key motion, which correspond to significant motion snapshots aligned with phoneme changes in the audio. Once the key motion indexes are determined, a key motion decoder interprets associated 3D facial meshes from corresponding audio features. This highlights those distinct facial expressions and facilitates lip-speech synchronization.
The cross-modal motion completion module expands non-continuous key motions into a full sequence of continuous face motions using audio features as guidance. This process enhances audio-mesh alignment and improves the temporal smoothness of output facial mesh sequences.
The contributions of our work are summarized as follows:
\begin{itemize}
    \item We propose a progressive learning mechanism to generate speech-driven 3D talking faces. It uses linguistic priors to initially generate key motions, and then interpolate key motions into complete motions via data-driven priors.
    \item We propose the use of phoneme-based localization methods to capture key facial motions. It effectively captures significant expression transitions aligned with phoneme changes in audio, improving lip-speech synchronization. 
    \item 
    We design a cross-modal facial motion completion module to produce full sequences of 3D talking faces using synthesized key motions and audio features. It further enhances lip-speech synchronization accuracy while facilitating temporal coherence in facial motions.
\end{itemize} 
Extensive experimental comparisons on the BIWI~\cite{fanelli20103} and VOCASET~\cite{cudeiro2019capture} datasets demonstrate that our method outperforms existing state-of-the-art approaches in more accurate and realistic talking face generation. Detailed ablation studies confirm the effectiveness of our proposed key motion capture technique. Additionally, consistent improvements in results by combining our proposed scheme with existing methods validate the efficacy of our design.
\section{Related Work}
\label{sec:related}
%
%
%
While existing research \cite{alghamdi2022talking, chen2020talking, chung2017out, das2020speech, guo2021ad, ji2022eamm, ji2021audio, liang2022expressive, liu2022semantic, pang2023dpe, prajwal2020lip, shen2022learning, vougioukas2020realistic, wang2022one, yi2020audio, zhang2023metaportrait, zhou2021pose}  focuses on 2D talking heads, 
we focus on audio-driven 3D facial animations in this work, which can be roughly categorized into linguistics-based and data-driven methods.
%
%

%
\subsection{Linguistic-based Methods}
Linguistic-based methods \cite{edwards2016jali, massaro201212, taylor2012dynamic, xu2013practical, bao2023learning,fisher1968confusions,lewis1991automated, cohen2001animated, mattheyses2015audiovisual} establish a set of intricate phoneme-to-viseme mapping rules for animating the mouth.
For example, the dynamic viseme model proposed by Taylor $et \ al.$
 \cite{taylor2012dynamic} exploits the one-to-many mapping of phonemes to lip motions.
JALI \cite{edwards2016jali} considers the many-to-many mapping between phonemes and visemes.
%
More recently, Bao $et \ al.$ \cite{bao2023learning} introduced a novel parameterized viseme fitting algorithm that extracts viseme parameters from speech videos using phonemic priors.
Leveraging linguistic priors, these methods indicate the articulation process by providing animators with explicit control over animation, thus boosting their performance in lip-speech synchronization. However, these approaches primarily focus on animating the lip region, lacking a comprehensive strategy for animating the entire face. In our work, we leverage linguistic priors to detect key frames with significant expression changes from audio in an analytical manner, without the need for supervised training.
%
%
\subsection{Data-driven Methods}
With the development of deep learning technology and the availability of high-quality datasets, data-driven methods \cite{cao2005expressive, cudeiro2019capture, fan2022faceformer, habibie2021learning, karras2017audio, liu2015video, pham2018end, taylor2017deep, wang20213d, xing2023codetalker, peng2023selftalk, richard2021meshtalk, wu2023speech, nocentini2023learning, zhang20213d, tang2024dphms} were proposed to synthesize entire 3D facial animation.
Some methods \cite{karras2017audio, cudeiro2019capture, richard2021audio, fan2022faceformer, wu2023speech} attempt to establish a direct audio-to-visual mapping through regression.
Person-specific approaches \cite{karras2017audio,richard2021audio} can usually obtain plausible facial motions because of the relatively consistent talking style. VOCA \cite{cudeiro2019capture} incorporates a robust audio feature extraction model capable of capturing various speaking styles, which can generate realistic speaker-independent animation and shows its wide applicability. MeshTalk~\cite{richard2021meshtalk} constructed a categorical latent space to adaptively generate motions based on the separated audio-correlated and audio-uncorrelated facial information. FaceFormer~\cite{fan2022faceformer} introduced two biased attention mechanisms and integrated the self-supervised pre-trained speech representations for the ill-posed and data scarcity issues. 
CodeTalker~\cite{xing2023codetalker} proposed the discrete motion prior which regards the cross-modal mapping as a code query task in a finite proxy space of the learned codebook. SelfTalk~\cite{peng2023selftalk} proposed a self-supervised approach to construct a lip-reading interpreter and speech recognizer to enhance the comprehensibility of generated lip movements.
%
%
While data-driven approaches have shown impressive performance, accurately learning cross-modal audio-visual mappings remains challenging due to inherent uncertainties. These methods often regress the entire sequence, leading to over-smoothing and a lack of detailed facial dynamics. In contrast, our approach employs a coarse-to-fine learning mechanism that separates the problem into key motion capture and motion completion stages. This approach effectively mitigates cross-modal mapping uncertainties and reduces learning complexity, resulting in more precise and dynamic facial animation synthesis.

\section{Method}
\begin{figure}[t]
    \centering
        \subfloat[The overview pipline of our proposed KMTalk.]{\label{fig:pipe1}
        \includegraphics[width = 1\linewidth]{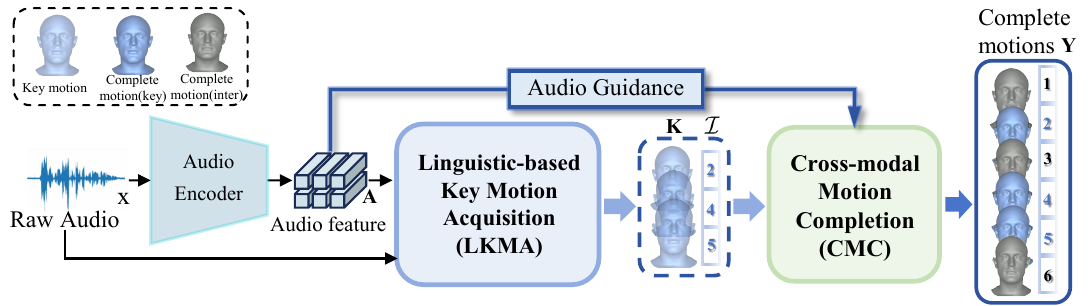}}
        \hfill
        \subfloat[The details of two key modules in KMTalk.]{\label{fig:pipe2}
        \includegraphics[width = 1\linewidth]{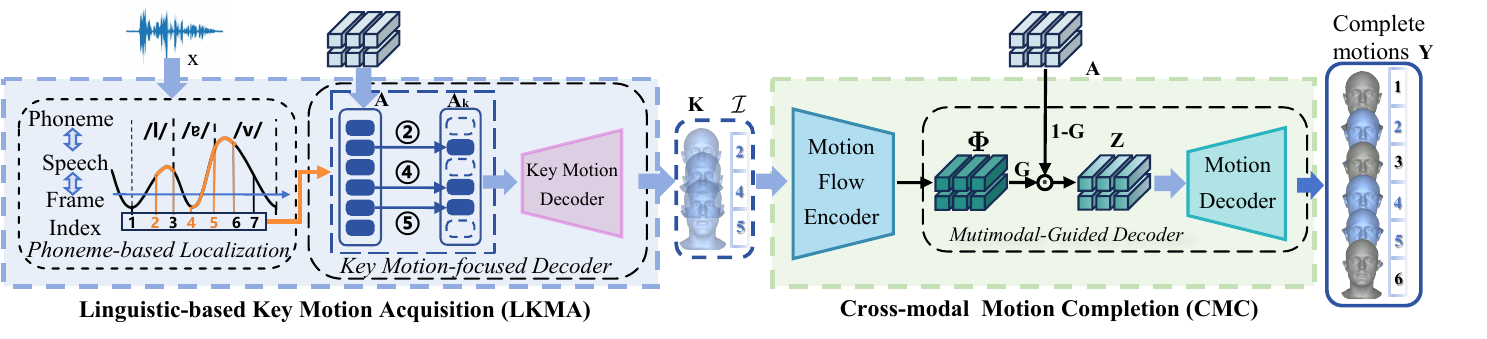}}
        \hfill

    \hfil
    \caption{
    Fig. \ref{fig:pipe1} illustrates the overview pipeline of our proposed KMTalk. Initially, the Audio Encoder takes the input raw audio $\mathbf{x}$ and encodes it into audio features $\mathbf{A}$. Subsequently, in the LKMA module, key motions $\mathbf{K}$ are generated from the audio $\mathbf{x}$ and $\mathbf{A}$. Finally, the CMC module reintroduces audio features $\mathbf{A}$ to extend these key motions $\mathbf{K}$ into a full sequence $\mathbf{Y}$. 
    Fig. \ref{fig:pipe2} presents the details of two key modules in KMTalk. In the Linguistic-based Key Motion Acquisition, a Phoneme-based Localization Method is used to identify key motion indices $\mathcal{I}$ from raw audio $\mathbf{x}$. Based on audio features $\mathbf{A}$ and $\mathcal{I}$, the Key Motion-focused Decoder generates key motions $\mathbf{K}$. In the Cross-modal Motion Completion, the Motion Flow Encoder processes $\mathbf{K}$ and $\mathcal{I}$, producing motion flow features $\mathbf{\Phi}$. Then, with the dynamic fusion weight $\mathbf{G}$, the Multimodal-Guided Decoder combines $\mathbf{\Phi}$ and $\mathbf{A}$ to decode the final motion sequence $\mathbf{Y}$.
    }
    \label{fig:framework}
\end{figure}
\subsection{Overview}
Let $\mathbf {x}$ represents the raw audio input and $ \mathbf{ \hat{Y}}=({\mathbf{\hat{y}}_1},...,{\mathbf{\hat{y}}_N}) \in \mathbbm{R}^{N\times V\times 3}$ denotes the corresponding ground-truth sequence of facial movement over a neutral template, where $N$ indicates the number of visual frames and $V$ denotes the number of vertices in the facial mesh.
The objective is to synthesize $\mathbf{Y}=({\mathbf{y}_1},...,{\mathbf{y}_N})$ that is similar to $\mathbf{\hat{Y}}$, driven by the raw audio $\mathbf {x}$.
The generated sequence should ensure lip synchronization with the audio while exhibiting natural facial movements.

Due to the domain gap between modalities and the ill-posed nature of directly translating audio to facial movement sequences, it is a challenging task that often results in over-smooth or poorly synchronized lip movements.
To address these issues, this paper introduces a coarse-to-fine approach with key motion embedding, integrating both linguistic and data-driven priors. 
The overview pipeline is presented in Fig. \ref{fig:pipe1}.
In the LKMA module, linguistic priors are introduced to locate and generate higher-quality key motions (see Sec. \ref{sec:KMA}), followed by the CMC module, where these key motions are fleshed out into a complete sequence of facial motions (see Sec. \ref{sec:MC}).
\subsection{Linguistic-based Key Motion Acquisition}
\label{sec:KMA}
In the realm of audio-driven 3D facial animation, it presents significant challenges to precisely define which frames constitute key motions. 
An alternative and simple solution is to use uniform or random sampling to determine the positions of these key motions.
Although these approaches can boost performance to a certain degree due to the reduced learning complexity (refer to Sec. \ref{sec:ablation} for ablation studies), 
they fail to utilize the correlation between audio content and facial movements. 
%
%
However, we can leverage linguistic priors to capture pronounced articulatory actions, which are identifiable at phoneme boundaries. This approach circumvents the issue of over-smooth in the output sequence, thereby enhancing the overall quality of the results.
%
%

As shown in the left of Fig. \ref{fig:pipe2}, the Linguistic-based Key Motion Acquisition (LKMA) module receives as inputs the raw audio $\mathbf{x}$ and the audio features $\mathbf{A}=(\mathbf{a}_1,...,\mathbf{a}_N)\in \mathbbm{R}^{N\times d}$, where the audio features are derived from the Audio Encoder that utilizes the wav2vec 2.0 pre-trained model \cite{baevski2020wav2vec}, processing the raw audio $\mathbf{x}$ as its input.
Then it takes raw audio $\mathbf {x}$ as input to produce the key motion indices $\mathcal{I}=\{i_1,...,i_m\}$, where $i_j \in \{1,..., N\}, \forall i_j \in \mathcal{I}$, through the proposed Phoneme-based Localization method. 
Subsequently, the Key Motion-focused Decoder utilizes the audio features $\mathbf{A}$ and the key motion indices $\mathcal{I}$ to generate key motions $\mathbf{K}=(\mathbf{k}_{i_1},...,\mathbf{k}_{i_m})\in \mathbbm{R}^{m\times V \times 3}$ consisting of $m$ frames of facial movement which are located on the key motion positions, where $\mathbf{k}_{i_j} \simeq  \mathbf{\hat{y}}_{i_j}, \forall i_j \in \mathcal{I}$.
The process of the LKMA module is expressed as:
\begin{equation} \label{eq:SSG} 
\mathcal{I},\mathbf{K} = {\rm LKMA}(\mathbf{x}, \mathbf{A}).
\end{equation}
%
%
%

\noindent \textbf{Phoneme-based Localization.}
%
%
At phoneme boundaries, a noticeable offset is observed in the articulator movement, with visualization results available in Appendix~\ref{sec:phoneme visualization}. 
Furthermore, the phoneme boundary effects underscore the ease with which the boundaries of phonemes can be perceived \cite{iverson2000perceptual}.
Both experimental and theoretical analyses have demonstrated a distinct position-mapping relationship between the phoneme boundaries in the audio and the significant elements in the motion sequence, specifically the key motions.
%
%
%
The mapping relationship can be harnessed to facilitate the initial alignment between the audio and visual modalities.
%
%

Specifically, an Automatic Speech Recognition model \cite{ma2303auto} is first utilized to obtain the text content from the raw audio $\mathbf {x}$. 
Then, a Montreal Forced Aligner \cite{mfa_english_mfa_acoustic_2023} is adopted to align the audio and the text, producing the start and the end timestamps for each phoneme.
Finally, the indices of these motion frames corresponding to the timestamps are regarded as key motion indices $\mathcal{I}$, and the corresponding motions compose the key motions $\mathbf{K}$.

\noindent \textbf{Key Motion-focused Decoder.}
\label{sec:KMG Net}
%
%
%
%
%
%
%
It is utilized to synthesize key motions $\mathbf{K}$ of superior quality. 
Initially, employing $\mathcal{I}$ as indices, we extract the corresponding aligned audio features $\mathbf{A_k} = (\mathbf{a}_{i_1},...,\mathbf{a}_{i_m}) \in \mathbbm{R}^{m\times d}$ from the comprehensive audio features $\mathbf{A}$.
%
Subsequently, it adopts a modified transformer-based architecture, processes $\mathbf{A_k}$ to generate the key motions $\mathbf{K}$.
%

\noindent \textbf{Loss Function.}
Intuitively, a straightforward approach to optimize the key motions $\mathbf{K}$ involves utilizing $\mathcal{I}$ to index $\mathbf{\hat{Y}}$, resulting in $\mathbf{\hat{Y}_k} = (\mathbf{\hat{y}}_{i_1}, \ldots, \mathbf{\hat{y}}_{i_m})$, which serves as the supervision for the training process.
%
%
However, key motions typically occupy non-adjacent positions within the entire sequence.
%
%
Hence, given the lack of inter-frame contextual information, attempting direct frame-by-frame regression of $\mathbf{K}$ towards $\mathbf{\hat{Y}_k}$ may fall short in achieving accurate facial expressions, as well as producing smooth and realistic animation.
%
%

To address this limitation, we adopt a pseudo-complete sequence training method that utilizes the ground-truth frame labels $\mathbf{\hat{Y}}$ at the non-key indices $\mathcal{I'}=\{1,...,N\}\setminus\mathcal{I}$ and the generated key motions $\mathbf{K}$ at the key indices $\mathcal{I}$ to form a predicted \textbf{p}seudo-complete sequence $\mathbf{Y_{p}} \in \mathbbm{R}^{N\times V \times 3}$.
%
%
%
Then, the model is trained by minimizing the loss between the pseudo-complete sequence $\mathbf{Y_{p}}$ and the ground-truth sequence $\mathbf{\hat{Y}}$.
This enables the model to capture subtle changes between key motions and adjacent ground-truth frames, thereby mitigating inter-frame jitter and achieving more accurate regression of facial expressions.
%
%
Following the SelfTalk \cite{peng2023selftalk}, the loss function is formulated as:
\begin{equation} \label{eq:loss1} 
    \mathcal{L}_{LKMA} = \lambda_1 \mathcal{L}_{rec} + \lambda_2 \mathcal{L}_{vel} + \lambda_3 \mathcal{L}_{lat} + \lambda_4 \mathcal{L}_{ctc},
\end{equation}
where $\lambda_1=1000.0, \lambda_2=1000.0, \lambda_3 = 0.001$, and $\lambda_4=0.0001$ in all of our experiments.
$\mathcal{L}_{rec}$, $\mathcal{L}_{vel}$, and  $\mathcal{L}_{lat}$ are measured by mean square error, while $\mathcal{L}_{ctc}$ is quantified by CTC Loss.
The reconstruction loss $\mathcal{L}_{rec}$ quantifies the discrepancy between the predicted and the ground-truth facial movements. 
The velocity loss $\mathcal{L}_{vel}$ reduces frame jitter, ensuring smooth and natural lip movements.
The latent consistency loss $\mathcal{L}_{lat}$ assesses the variance between latent features extracted from both the audio and lip shape encoders, aiming to align the learned audio and lip features. 
Lastly, the text consistency loss $\mathcal{L}_{ctc}$ evaluates the difference between the lip-reading decoder's output and the original text, ensuring the intelligibility of the lip-reading results.
%
%
%
%
%
\subsection{Cross-modal Motion Completion}
\label{sec:MC}

A straightforward method to obtain a complete talking face sequence is to directly integrate key motions into the entire face movements. However, it is important to note that while key motions capture essential facial dynamics, they may not encompass all details of non-key motions. This direct integration could result in mismatches between augmented non-key motions and their corresponding audio segments (see Sec. \ref{sec:ablation} for ablation studies).
To mitigate this issue, we introduce a Cross-modal Motion Completion (CMC) module that jointly combines the audio features $\mathbf{A}$, key motions $\mathbf{K}$, and key motion indices $\mathcal{I}$ to generate a complete sequence of 3D facial meshes $\mathbf{Y}$. The process can be formulated as:
\begin{equation} \label{eq:CSS} 
    \mathbf{Y} = {\rm CMC}(\mathbf{A, K},\mathcal{I}).
\end{equation}
The details of the CMC module are illustrated on the right of Fig. \ref{fig:pipe2}.

\noindent \textbf{Motion Flow Encoder.}
Key motions serve as a kinematic prior for the remaining frames, offering valuable insights into facial dynamics. 
To effectively capture the motion flow information provided by key motions, we draw inspiration from some manifold methods \cite{mo2023continuous} to acquire motion flow features $\mathbf{\Phi} \in \mathbbm{R}^{N\times d}$ from the key motions $\mathbf{K}$.
%
%
Specifically, we first encode the key motions $\mathbf{K}$ into key motion context tokens $\mathbf{\Phi_{k}} \in \mathbbm{R}^{m\times d}$ by multiple transformer-encoder layers.
At the same time, the indices of non-key motions $\mathcal{I'}$ are encoded into positional encodings by a sinusoidal position embedding layer, representing the non-key frame positions. 
Then, we adopt the cross-attention layers to extract the intermediate tokens $\mathbf{\Phi_{non\text{-}key}} \in \mathbbm{R}^{(N-m)\times d}$, with key and value from linear transformations of $\mathbf{\Phi_{k}}$ and the query is the positional encodings of non-key frames indices. 
%
%
Above all, the implicit motion manifold proposed in CITL \cite{mo2023continuous} is utilized to arrange $\mathbf{\Phi_{k}}$ and $\mathbf{\Phi_{non\text{-}key}}$ based on their indices, followed by a 1D convolution for fusion, ultimately obtaining the motion flow features $\mathbf{\Phi}$.
%
%
%
%

\noindent \textbf{Multimodal-Guided Decoder.}
The non-key motions' features in the complete motion sequence feature estimation are derived from the global context interpolation of key motions, which has a certain degree of information loss due to the audio feature selection process in the Key Motion-focused Decoder (Sec. \ref{sec:KMG Net}).
Hence, we have devised a multimodal-guided decoding approach that incorporates audio modalities to furnish comprehensive information across the entire temporal scale, alongside motion flow to offer facial motion priors. These elements serve to guide and constrain the decoding process, thereby facilitating the precise generation of the motion sequence.
Technically, we simply employ the gated mechanism \cite{yu2020towards, yue2020robustscanner,dai2023disentangling} for the modality fusion, which can be formulated as:
\begin{equation} \label{eq:fusion1} 
\mathbf{G} = \sigma([\mathbf{A},\mathbf{\Phi}]\mathbf{W}),
\end{equation}
\begin{equation} \label{eq:fusion2} 
\mathbf{Z} = \mathbf{G} \odot \mathbf{A} + (1-\mathbf{G})\odot \mathbf{\Phi},
\end{equation}
where $\odot$ represents the element-wise multiplication operation, $[\cdot,\cdot]$ denotes the concatenation operation,  $\mathbf{W}\in \mathbbm{R}^{2d\times d}$ is a parameter and $\sigma$ is a sigmoid function, while $\mathbf{G}\in \mathbbm{R}^{N\times d}$ dynamically selects features from the audio features $\mathbf{A}$ and the motion flow features $\mathbf{\Phi}$.
%
%
Then, the Motion Decoder, a transformer-based structural model, is employed to transform the fusion features $\mathbf{Z}$ into the complete 3D facial movement sequence $\mathbf{\hat{Y}}$.
%
%

\noindent \textbf{Loss Function.}
We train the CMC module utilizing the reconstruction loss and velocity loss.
The total loss function is defined as:
\begin{equation} \label{eq:loss2} 
\begin{split}
    \mathcal{L}_{CMC} = \mathcal{L}_{rec} + \mathcal{L}_{vel} .
\end{split}
\end{equation}
%
\section{Experiment}

\begin{table*}[t]
\caption{
%
%
Quantitative comparisons on the BIWI-Test-A and VOCA-Test datasets. The results of Lip-Vertex Error (LVE) and the upper-Face Dynamics Deviation (FDD) are reported. For both metrics, the lower the better.
}
\label{tab:eval}
\resizebox{\columnwidth}{!}{%
\setlength{\tabcolsep}{5mm}{
\begin{tabular}{ccccccc}
\toprule
\multirow{3}{*}{Methods}              & \multicolumn{2}{c}{BIWI-Test-A}                     &  & \multicolumn{2}{c}{VOCA-Test}                       \\ \cline{2-3} \cline{5-6} 
                                                              & LVE$\downarrow$          & FDD$\downarrow$          &  & LVE$\downarrow$          & FDD$\downarrow$          \\
                                                              & $\times 10^{-4}{\rm mm}$ & $\times 10^{-5}{\rm mm}$ &  & $\times 10^{-5}{\rm mm}$ & $\times 10^{-7}{\rm mm}$ \\ \hline
VOCA \cite{cudeiro2019capture}                     & 6.5563                   & 8.1816                   &  & 4.9245                   & 4.8447                   \\
MeshTalk \cite{richard2021meshtalk}                 & 5.9181                   & 5.1025                   &  & 4.5441                   & 5.2062                   \\
FaceFormer \cite{fan2022faceformer}                 & 5.3077                   & 4.6408                   &  & 4.1090                   & 4.6675                   \\
CodeTalker \cite{xing2023codetalker}                & 4.7914                   & 4.1170                   &  & 3.9445                   & 4.5422                   \\
SelfTalk \cite{peng2023selftalk}                  & 4.2485                   & 3.5761                   &  & 3.2238                   & 4.0912                   \\
\textbf{KMTalk (Ours)}                                    & \textbf{3.9654}          & \textbf{2.5446}          &  & \textbf{2.2639}          & \textbf{4.0594}          \\ \bottomrule
\end{tabular}%
}}
\end{table*}
\noindent  \textbf{Dataset.}
\textit{BIWI}~\cite{fanelli20103} consists of 40 paired audio-visual sentences from 14 subjects.
The 3D facial geometries, consisting of 23370 vertices, were captured at a frame rate of 25fps, and the average duration of each sequence was 4.67 seconds.
We adopt the same evaluation protocol as FaceFormer \cite{fan2022faceformer} on the BIWI dataset.
Specifically, the training set (BIWI-Train) comprises 190 sentences, while the validation set (BIWI-Val) encompasses 24 sentences.
The dataset is divided into two testing sets: BIWI-Test-A, which comprises 24 sentences articulated by six subjects observed during training, and BIWI-Test-B, which consists of 32 sentences uttered by eight unseen subjects.
\textit{VOCASET}~\cite{cudeiro2019capture} consists of 480 paired audio-visual sequences from 12 subjects.
Each sequence is recorded at a frame rate of 60fps and ranges in duration from 3 to 4 seconds.
The 3D face mesh for each sequence consists of 5023 vertices.
To ensure a fair comparison, we used identical training (VOCA-Train), validation (VOCA-Val), and testing (VOCA-Test) partitions as methods\cite{fan2022faceformer,xing2023codetalker,peng2023selftalk}.

\noindent \textbf{Baselines.}
We compare against current state-of-the-arts method, including 

\noindent VOCA~\cite{cudeiro2019capture}, MeshTalk~\cite{richard2021meshtalk}. FaceFormer \cite{fan2022faceformer}, CodeTalker \cite{xing2023codetalker}, and SelfTalk \cite{peng2023selftalk}. Faceformer~\cite{fan2022faceformer}  employs a transformer-based model to incorporate long-term audio context and synthesizes sequential motions in an autoregressive manner. CodeTalker~\cite{xing2023codetalker} introduces discrete motion priors to enable self-reconstruction of real facial movements, mitigating the issue of excessive smoothing in facial motion. SelfTalk~\cite{peng2023selftalk} designs a learning-based recognizer to minimize the domain gap between diverse modalities.

\noindent  \textbf{Evaluation Metrics.}
    Following CodeTalker \cite{xing2023codetalker} and SelfTalk \cite{peng2023selftalk}, we adopt two metrics for the quantitative evaluation of speech-driven facial animation: \textit{lip vertex error} (LVE) to measure lip synchronization and \textit{upper-face dynamics deviation} (FDD) to assess the overall facial dynamics.
The LVE for each frame is defined as the maximal L2 error among all lip vertices for each frame and takes the average over all frames.
This L2 error is computed by comparing the predictions with the processed 3D face geometry data.
FDD is introduced to quantify the variation in facial dynamics between a synthetic motion sequence and the reference sequence.
The implementation of FDD is to calculate the difference between the variances of vertex offsets in the upper-face region and the variances of ground truth vertex offsets. In addition, we visualize the prediction results for qualitative evaluation.\par
\noindent\textbf{Implementation Details.}
For a fair comparison, KMTalk operates at a frame rate of 30 fps on VOCASET and 25 fps on BIWI, following the setting of previous methods~\cite{fan2022faceformer,xing2023codetalker,peng2023selftalk}. Also, it can naturally adapt to a higher frame rate, as shown in the Appendix~\ref{sec:frame rate}. In the LKMA module, we first employ the Phoneme-based Localization method to process the raw audio and obtain key motion indices for data preprocessing, which costs less than 10 minutes on two datasets~\cite{cudeiro2019capture,fanelli20103}.
Secondly, we train the Key Motion Decoder on a single NVIDIA RTX 3090 for 200 epochs (about 2 hours) using the Adam optimizer \cite{kingma2014adam}. The learning rate is initialized as $10^{-4}$, and the mini-batch size is set to 1. 
In the CMC module, we train for 200 epochs (approximately 2 hours) with the same training settings as the Key Motion Decoder.
It is noteworthy that, since the training of the two modules is independent of each other, we can train both modules concurrently to enhance training efficiency.
\subsection{Comparisons against State-of-the-art Methods}
\begin{figure*}[t]
    \centering
    \includegraphics[width=\linewidth]{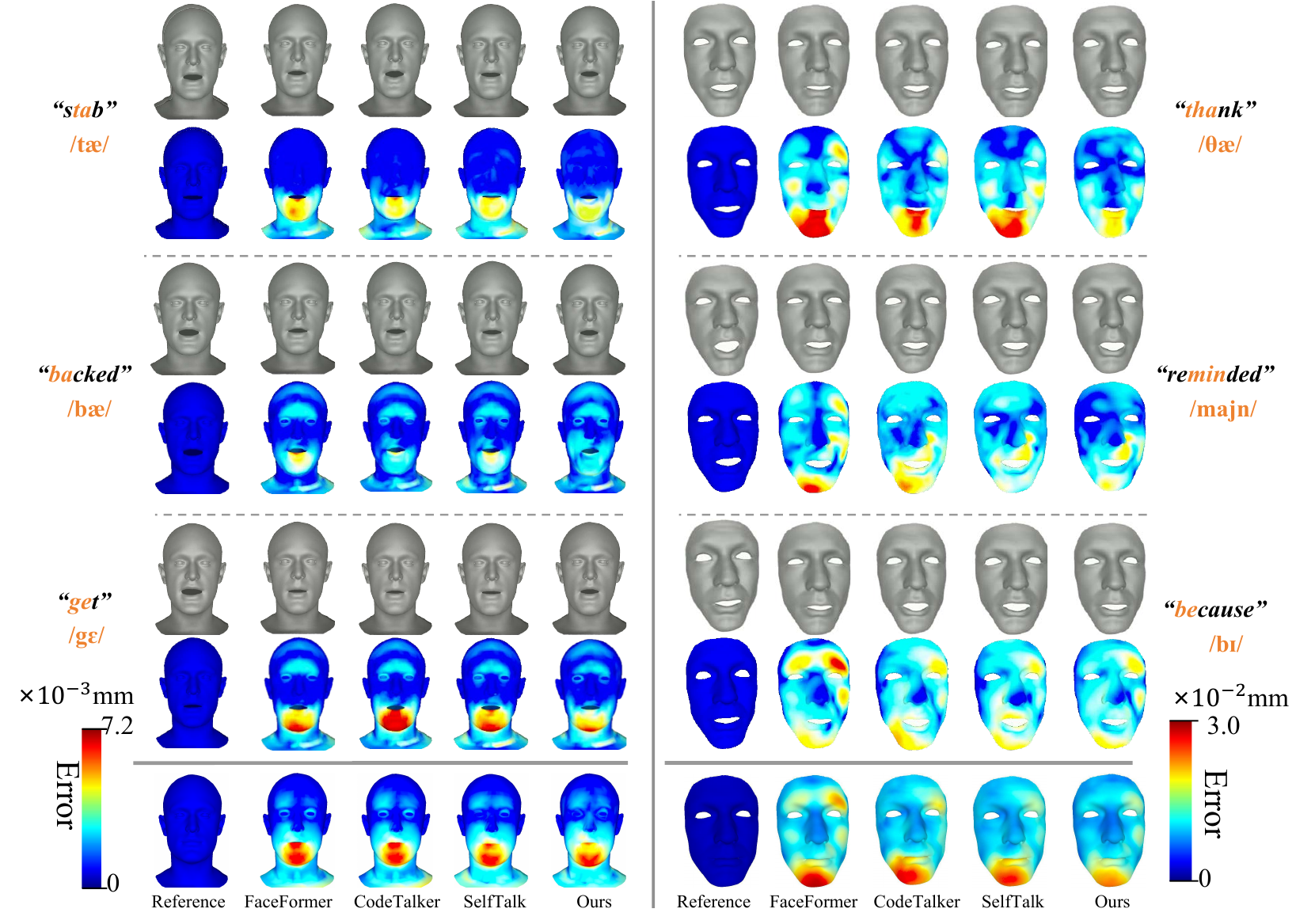}
    \caption{Qualitative comparisons on VOCA-Test (left) and BIWI-Test-B (right).
    %
    We provide visual comparisons of facial animations synchronized with six syllables extracted from the test speech sequences.
    The 1st, 3rd, and 5th rows display synthesized meshes and their corresponding ground-truths, while the 2nd, 4th, and 6th rows visualize the L2 loss for individual frames.
    Our method demonstrates more precise mouth movement on syllables like /\ae/ that require a wide-open mouth. For syllables that start with a closed mouth and then slightly open, such as /b\textipa{I}/, our KMTalk generates more synchronized motion sequences visually.
    The last row visualizes the mean square errors of different methods across all sentences in the test set for a specific subject.
    }
    \label{fig:baseline}
\end{figure*}
\noindent\textbf{Quantitative Comparisons.}
%
We computed the lip vertex error (LVE) and facial dynamics deviation (FDD) for all sequences within the BIWI-Test-A and VOCA-Test datasets.
According to Table \ref{tab:eval}, our proposed KMTalk demonstrated lower errors compared to the alternative methods examined.
Notably, the lip vertex error for our method on the VOCA-Test dataset is 30\% lower than the recently introduced SelfTalk \cite{peng2023selftalk}, and the FDD is 27\% lower than SelfTalk on the BIWI-Test-A dataset, providing compelling evidence for the advantages of our proposed KMTalk.
This indicates that our approach is more effective in achieving audio-visual alignment, thereby leading to improved lip synchronization.

\noindent  \textbf{Qualitative Comparisons.}
%
%
In Fig. \ref{fig:baseline}, we visualize the output facial meshes from different methods as well as ground truths for reference. Additionally, we display error maps calculated from the vertex L2 loss between the generated and ground truth meshes. It is evident that our method consistently yields lower errors across different speech sequences, demonstrating its ability to generate more accurate facial animation sequences.
Notably, for representative syllables (\textit{e.g.} /\ae/), KMTalk closely approximates ground truth, excelling in synthesizing accurate lip movements for syllables requiring significant mouth opening. Additionally, for syllables starting with mouth closure followed by a slight opening (e.g., /b\textipa{I}/), KMTalk produces more natural and synchronized motions. 
We recommend that readers watch the supplementary video for more detailed comparisons. It showcases KMTalk's capability to generate coherent, realistic animations with precise lip synchronization.

\noindent  \textbf{User Studies.}
A user study stands as a dependable evaluation method in the context of 3D talking faces. 
Following the strategy of Faceformer~\cite{fan2022faceformer}, we conduct pairwise comparisons between our method and baselines~\cite{fan2022faceformer, xing2023codetalker, peng2023selftalk}, as well as ground truths.
This study encompassed the assessment of two key metrics: perceptual lip synchronization and facial realism. 
Participants were presented with side-by-side comparisons and were tasked with selecting the better facial animation based on their personal preferences.
We computed the ratio of user preferences as a measurement of satisfaction evaluation on BIWI-Test-B and VOCA-Test. 
We randomly sampled 30 examples from each test set and compared the performance of KMTalk with four aforementioned settings on each sample.
Therefore, we constructed a total of 240 different video pairs and randomly selected 24 video pairs for the two metrics assessments for each participant.
Our user study involved 30 participants with a strong capability for audio-visual perception, resulting in 720 effective evaluation entries.
As demonstrated in Table \ref{tab:user-study}, our approach indicates superior perceptual lip synchronization and facial realism. 
For instance, a noteworthy 60.0\% of users favored our lip synchronization method on BIWI-Test-B in comparison to SelfTalk \cite{peng2023selftalk}. 
Overall, it shows that KMTalk can generate more favorable facial animations from speeches.
\begin{table}[t]
\centering
\caption{User study results on BIWI-Test-B and VOCA-Test.}
\label{tab:user-study}
\resizebox{0.9\columnwidth}{!}{%
\renewcommand\arraystretch{1.1}
    {
    \scalebox{0.9}{
    \begin{tabular}{cccccc}
\hline
\toprule
\multirow{2}{*}{Method}                       & \multirow{2}{*}{Metric} & \multicolumn{2}{c}{BIWI-Test-B} & \multicolumn{2}{c}{VOCA-Test} \\
                                              &                         & competitor   & ours             & competitor  & ours            \\ \hline
\multirow{2}{*}{\textbf{Ours vs. FaceFormer}} & Lip Sync                & 24.4\%       & \textbf{75.6\%}  & 26.7\%      & \textbf{73.3\%} \\
                                              & Realism                 & 25.6\%       & \textbf{74.4\%}  & 28.9\%      & \textbf{71.1\%} \\ \hline
\multirow{2}{*}{\textbf{Ours vs. CodeTalker}} & Lip Sync                & 31.1\%       & \textbf{68.9\%}  & 37.8\%      & \textbf{62.2\%} \\
                                              & Realism                 & 27.8\%       & \textbf{72.2\%}  & 35.6\%      & \textbf{64.4\%} \\ \hline
\multirow{2}{*}{\textbf{Ours vs. SelfTalk}}   & Lip Sync                & 40.0\%       & \textbf{60.0\%}  & 43.3\%      & \textbf{56.7\%} \\
                                              & Realism                 & 38.9\%       & \textbf{61.1\%}  & 41.1\%      & \textbf{58.9\%} \\ \hline
\multirow{2}{*}{\textbf{Ours vs. GT}}         & Lip Sync                & 54.4\%       & \textbf{45.6\%}  & 56.7\%      & \textbf{43.3\%} \\
                                              & Realism                 & 52.2\%       & \textbf{47.8\%}  & 56.7\%      & \textbf{43.3\%} \\ \bottomrule \hline
\end{tabular}%
    } 
}}
\end{table}
\begin{figure}[t]
    \centering
    \includegraphics[width=\linewidth]{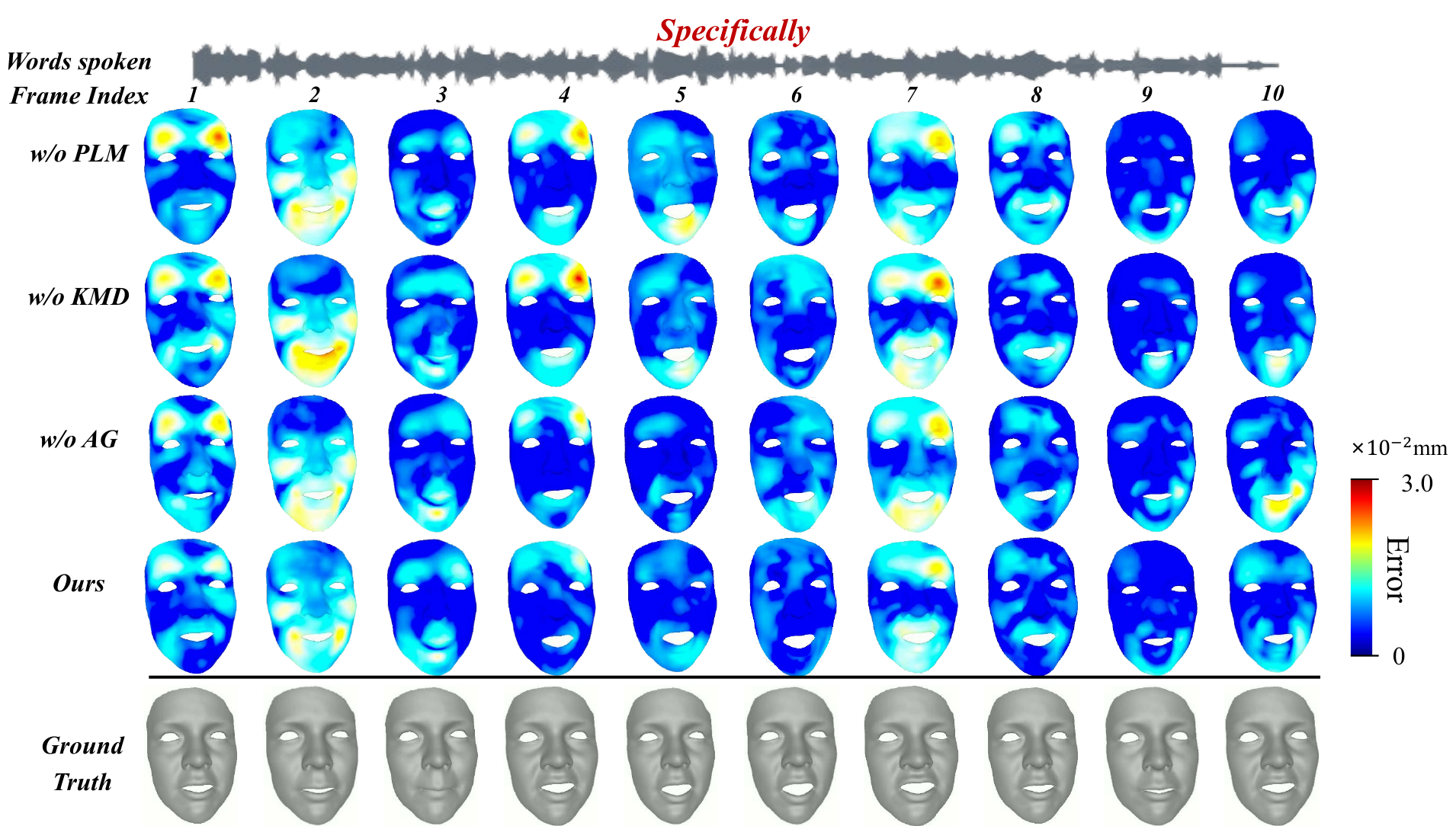}
    \caption{Qualitative ablation studies on the input speech ``specifically''. For each method variant, we removed one of three modules: PLM (Phoneme-based Localization Method), KMD (Key Motion-focused Decoder), and AG (Audio Guidance in CMC). Error maps between generated and the ground-truth mesh sequence were visualized. Our final model yielded the best results, showcasing the effectiveness of each module.
    }
    \label{fig:ablation studies}
\end{figure}
\vspace{0.5cm}
\begin{table}[t]
\centering
\caption{Ablation study for our components on BIWI-Test-A. 
}
\label{tab:ablation study}
\resizebox{0.8\columnwidth}{!}{%
\begin{tabular}{cccccc}
\hline
\toprule
\textbf{Phoneme-based}       & {\textbf{Key Motion-focused}} & \textbf{Audio Guidance} & \textbf{} & \multirow{2}{*}{\textbf{\textbf{LVE$\downarrow$}}} & \multirow{2}{*}{\textbf{\textbf{FDD$\downarrow$}}} \\ 
\textbf{Localization Method} & \textbf{Decoder}                                  & \textbf{in CMC}          &           &                                                    &                                                    \\ \hline 
---                   & ---                      & ---                 &           & 4.2485                                             & 3.5761                                             \\ 
                                             
---                 & \Checkmark                        & \Checkmark                 &           & 4.1648                                             & 2.8713                                             \\
\Checkmark                   & ---                      & \Checkmark                 &           & 4.1381                                             & 2.9546                                             \\
\Checkmark                   & \Checkmark                        & ---               &           & 4.8859                                             & 3.2780                                             \\ \Checkmark                   & \Checkmark                        & \Checkmark                 &           & \textbf{3.9654}                                             & \textbf{2.5446} \\ \bottomrule \hline 
\end{tabular}%
}
\end{table}

\subsection{Ablation Studies}
\label{sec:ablation}
In this section, we perform ablation studies on to evaluate the influence of different components within our proposed KMTalk framework on the quality of the generated 3D talking faces.
%
%
The quantitative results on BIWI are in Table \ref{tab:ablation study}, and the qualitative results are in Fig.~\ref{fig:ablation studies}. 
In addition, the results of the ablation study on VOCA-Test can be found in the Appendix~\ref{sec:additional results}.
%
In Table \ref{tab:phomeme-robustic}, we further investigate the robustness of our approach to different Phoneme-based localization methods and the possible errors during phoneme extraction.
%

\noindent \textbf{What's the effect of the Phoneme-based localization method for key motion capture?} 
The Phoneme-based Localization Method enables us to identify key frames of speech with notable facial expression transitions. 
We can replace it with uniform sampling, neglecting the crucial content information of the audio.
Specifically, we experimented using uniform sampling at a rate of 33\% to closely align with the number of key motions. 
For further comparisons under different numbers of sampled elements, please refer to the Appendix~\ref{sec:key motion quantity}.
In Table \ref{tab:ablation study}, we observe degradation in all metrics with uniform sampling. 
This underscores the importance of accurate key motion capture in speech-driven talking face generation. Additionally, it showcases that the linguistic-based key motion capture is better equipped to mitigate audio-visual uncertainty and recover more precise facial motions.

\noindent \textbf{What's the effect of the Key Motion Decoder?}
%
%
Our method employs a specialized key motion decoder to generate facial meshes based on keyframe indices obtained from phoneme-based localization methods. An alternative approach to generating key motions is to select corresponding facial meshes from a complete motion sequence produced by an existing method such as SelfTalk \cite{peng2023selftalk}.
%
%
%
Table \ref{tab:ablation study} demonstrates that both metrics deteriorated but still outperformed the state-of-the-art method SelfTalk. This indicates that the Key Motion Decoder can enhance the quality of key motion generation, resulting in more plausible facial animations. It also suggests that even if the captured key motions are not accurate enough, the CMC module can further refine output full motions.

\noindent \textbf{What's the effect of audio guidance in the Cross-modal Motion Completion?}
The Cross-modal Motion Completion leverages audio features to guide the completion of the full motion sequence.  To assess its usefulness, we implement a method variant that removes the audio feature guidance. Table \ref{tab:ablation study} demonstrates a notable degradation in both metrics, particularly with a 24\% increase in Lip Vertex Error (LVE) and a 27\% increase in upper-face dynamics deviation (FDD).
%
%
This suggests that audio information plays a crucial role in refining fine-grained lip movements and enhancing audio-visual consistency and temporal smoothness.
Despite the degradation, the FDD metric can still outperform state-of-the-art methods, underscoring the significance of key motion capture for achieving temporally coherent full motion synthesis.


\noindent \textbf{Is our method sensitive to different Phomene-based Localization methods?} 
 We experimented with different Automatic Speech Recognition (ASR) models, such as Auto-avsr~\cite{ma2303auto} and Whisper~\cite{radford2023robust}.
The results in the first three rows of Table \ref{tab:phomeme-robustic} show that our KMTalk method consistently maintained high performance, achieving at least a 21\% improvement in FDD regardless of the ASR model used~\cite{ma2303auto,radford2023robust}. This highlights the robustness of our approach across various ASR models.
Besides, to simulate phoneme localization deviations, we shifted all key motion indices extracted by Auto-avsr~\cite{ma2303auto} one frame to the right. The results in the last row of Table~\ref{tab:phomeme-robustic} indicate negligible variations. This demonstrates the robustness of our method to inaccurate key frame localization. 
%
%

\hspace{-4mm}
\begin{minipage}[t]{0.4\textwidth}
\makeatletter\def\@captype{table}
\centering
\caption{Robust analysis of Phoneme-based Localization on BIWI-Test-A.}
\label{tab:phomeme-robustic}
    \renewcommand\arraystretch{1.5}
    {
    \scalebox{0.85}{
    \begin{tabular}{clcc}
\hline
\toprule
\multirow{2}{*}{Methods}        & LVE$\downarrow$ & FDD$\downarrow$ \\ 
            & \multicolumn{1}{l}{$\times 10^{-4}{\rm mm}$} & \multicolumn{1}{l}{$\times 10^{-5}{\rm mm}$} \\ \midrule
Auto-avsr~\cite{ma2303auto}           &  \bf 3.9654   &  \bf 2.5446   \\
Whisper-large~\cite{radford2023robust} &  4.0718   &  2.8141   \\
Whisper-tiny~\cite{radford2023robust}  &  4.0643   &   2.8083  \\  
Auto-avsr+offset           &  3.9991   & 2.6420     \\
\bottomrule\hline
\end{tabular}%
    } 
}
\end{minipage}
\hspace{5mm}
\begin{minipage}[t]{0.5\textwidth}
\makeatletter\def\@captype{table}
\centering
\caption{The results of integrating our proposed KMTalk with existing methods on BIWI-Test-A.}
\label{tab:general}
    \renewcommand\arraystretch{1.3}
    {
    \scalebox{0.77}{
    \begin{tabular}{clcc}
\hline
\toprule
\multirow{2}{*}{Methods}                              &          & LVE$\downarrow$                              & FDD$\downarrow$                              \\
                                                      &          & \multicolumn{1}{l}{$\times 10^{-4}{\rm mm}$} & \multicolumn{1}{l}{$\times 10^{-5}{\rm mm}$} \\ \hline
\multirow{2}{*}{FaceFormer \cite{fan2022faceformer}}  & Original & 5.3077                                       & 4.6408                                       \\
                                                      & After    & \textbf{5.2793}                              & \textbf{4.2654}                              \\ \hline
\multirow{2}{*}{CodeTalker \cite{xing2023codetalker}} & Original & 4.7914                                       & 4.1170                                       \\
                                                      & After    & \textbf{4.5096}                              & \textbf{3.9043}                              \\ \hline
\multirow{2}{*}{SelfTalk \cite{peng2023selftalk}}     & Original & 4.2485                                       & 3.5761                                       \\
                                                      & After    & \textbf{4.1122}                              & \textbf{2.8668}                              \\ \bottomrule \hline
\end{tabular}%
    } 
}
\end{minipage}
\subsection{Integration with Existing Methods}
\label{subsec: gene}
Existing approaches focus on enhancing prediction outcomes by designing elaborate priors or the learning-based recognizer, which may be highly coupled with the proposed architecture of these methods. Our KMTalk introduces a new learning strategy of speech-driven talking face generation, which is orthogonal to these approaches. Therefore, we can explore whether performance can be enhanced by applying our progressive learning scheme without the need for additional fine-tuning of their models.
%
Detailed implementation is in the Appendix~\ref{sec:Implementation Details}.
As shown in Table~\ref{tab:general}, the results of existing methods are improved after integration with our proposed progressive learning mechanism utilizing key motion embeddings. This further emphasizes the efficacy of our design.

\section{Conclusion}
%
%
In this work, we introduce KMTalk, a novel method for progressively learning 3D facial animation from speeches using key motion embeddings. It incorporates linguistic priors for key motion generation and extends them to a full motion sequence via data-driven priors. We propose phoneme-based localization methods to determine the temporal position of key facial motions, improving lip-speech synchronization by aligning motion transitions with phoneme changes. Additionally, we design a cross-modal facial motion completion module that synthesizes the entire motion sequence from key motions and audio features, enhancing lip-speech synchronization and motion coherence. Extensive evaluations of the datasets demonstrate KMTalk's superiority over existing methods, producing more accurate and realistic animations. Moreover, coupling our idea with existing methods consistently improves performance, further verifying the efficacy of our proposed progressive learning mechanism based on key motion acquisition. Although the proposed method has demonstrated its robustness to inaccurate keyframe localization, it may encounter errors in dialect variations. Integrating advanced ASR(Automatic Speech Recognition) technology in the future could enhance its adaptability to various speech patterns.
%


%
%
\noindent\textbf{Acknowledgement}
The research is partially supported by National Key Research and Development Program of China (2023YFC3502900), National Natural Science Foundation of China (No.62176093, 61673182), Key Realm Research and Development Program of Guangzhou (No.202206030001), Guangdong-Hong Kong-Macao Joint Innovation Project (No.2023A0505030016), Guangdong Natural Science Foundation (No. 2024A1515012217). 
\bibliographystyle{splncs04}
\bibliography{main}
\clearpage
\renewcommand\thesection{\Alph{section}}
\setcounter{page}{1}
\section*{Appendix}
\setcounter{section}{0}
%
%
\section{Overview}
In this supplementary material, we provide more implementation details on KMTalk (Sec.~\ref{sec:Implementation Details}), additional results and comparisons in Sec.~\ref{SecAddResult}, and more discussion (Sec.~\ref{sec:discussion}).
\section{Implementation Details}
\label{sec:Implementation Details}
\textbf{Network Architecture.} To enhance the reproducibility of our KMTalk approach, we provide the detailed network architecture for Linguistic-based Key Motion Acquisition ((Sec.~\textcolor{red}{3.2}) and Cross-modal Motion Completion (Sec.~\textcolor{red}{3.3}) in the main paper. The network architecture is presented in Table~\ref{tab:Parameter}. Our codebase will be released soon.
\begin{table}[ht]
\renewcommand{\arraystretch}{1.2}
\centering
\caption{Parameter illustration of network architectures. L($c_i,c_o$) denotes a linear layer with input channels of $c_i$ and output channels of $c_o$. Concat($v_1,v_2,c$) stands for the concatenation of $v_1$ and $v_2$ in dimension $c$. Sigmoid represents a sigmoid function. Weighted Sum($W$) denotes a weighted sum with the weight of $W$. TransformerDecoder($d\_model, nhead, dim\_ffd,num\_layers$) represents a transformer structure with the input channels $d\_model$, the number of heads in multi-head attention $nhead$, the channels of feedforward network $dim\_ffd$ and the number of decoder layers $num\_layers$.
PE($a$) is a position embedding layer where $a$ denotes the length of position vector.
MultiheadAttention($d\_model,nhead$) is an self-attention layer.
FFN($d\_model$) is a feed forward layer.
Conv1D represents 1D convolution operation.
The details of Manifold can be found in \cite{mo2023continuous}.}
\label{tab:Parameter}
\resizebox{\columnwidth}{!}{%
\begin{tabular}{c|l|l}
\hline
\toprule
Module                               & \multicolumn{1}{c|}{Input $\rightarrow$ Output}                      & \multicolumn{1}{c}{Layer Operation}                                                                                                                                                                                                                  \\ \hline
Audio Encoder                        & $\mathbf{x} \rightarrow \mathbf{A}(N,d)$                             & Wav2vec 2.0 pre-trained model \cite{baevski2020wav2vec}                                                                                                                                                                                                                   \\ \hline
Key Motion Decoder                   & $\mathbf{A_k}(m,d) \rightarrow \mathbf{K}(m,3\cdot V)$               & L($d,f$) $\rightarrow$ TransformerDecoder($f$,4,$2\cdot f$,1) $\rightarrow$ L($f,3\cdot V$)                                                                                                                                                          \\ \hline
\multirow{3}{*}{Motion Flow Encoder} & $\mathbf{K}(m,3\cdot V) \rightarrow \mathbf{\Phi_k}(m,d)$            & PE(16) $\rightarrow$ L(16+$f$,$f$) $\rightarrow$ {[}MultiheadAttention($f,8$) $\rightarrow$ FFN($f$){]}$\times 6$                                                                                                                                    \\ \cline{2-3} 
                                     & $\mathbf{T_k}(m) \rightarrow \mathbf{\Phi_{non\text{-}key}}(N-m,d)$             & PE(16) $\rightarrow$ L(16,$f$) $\rightarrow$ {[}MultiheadAttention($f,8$) $\rightarrow$ FFN($f$){]}$\times 6$                                                                                                                                        \\ \cline{2-3} 
                                     & $\mathbf{\Phi_k},\mathbf{\Phi_{non\text{-}key}} \rightarrow \mathbf{\Phi}(N,d)$ & \begin{tabular}[c]{@{}l@{}}Manifold(FFN($\mathbf{\Phi_k}$),$\mathbf{\Phi_{non\text{-}key}}$) $\rightarrow$ Conv1D $\rightarrow$ \\ {[}MultiheadAttention($f,8$) $\rightarrow$ FFN($f$){]}$\times 6$ \\ $\rightarrow$ Conv1D $\rightarrow$ L($f,d$)\end{tabular} \\ \hline
\multirow{3}{*}{Motion Decoder}      & $\mathbf{A}(N,d),\mathbf{\Phi}(N,d) \rightarrow \mathbf{W}(N,d)$     & Concat($\mathbf{A,\Phi}$,2) $\rightarrow$ L($2\cdot d, d$) $\rightarrow$ Sigmoid                                                                                                                                                                     \\ \cline{2-3} 
                                     & $\mathbf{A}(N,d),\mathbf{\Phi}(N,d) \rightarrow \mathbf{Z}(N,d)$     & Weighted Sum($\mathbf{W}$)                                                                                                                                                                                                                           \\ \cline{2-3} 
                                     & $\mathbf{Z}(N,d)\rightarrow \mathbf{Y}(N,3\cdot V)$            & L($d,f$) $\rightarrow$ TransformerDecoder($f$,4,$2\cdot f$,1) $\rightarrow$ L($f,3\cdot V$)                                                                                                                                                          \\ \bottomrule \hline 
\end{tabular}%
}
\end{table}

\noindent\textbf{Phoneme-based Localization Method.} The Phoneme-based Localization Method is proposed in this paper to locate the position of each phoneme. The specific procedure is as follows: First, the input speech signal is processed by an Automated Speech Recognition (ASR) module\cite{ma2303auto,radford2023robust}, which transcribes the speech into its corresponding textual representation based on acoustic and language models. Subsequently, the Montreal Forced Aligner (MFA) \footnote{Montreal Forced Aligner (MFA): \url{https://mfa-models.readthedocs.io/en/latest/mfa_phone_set.html}} module is employed to establish temporal alignments between the transcribed text and the original speech signal. This module utilizes advanced algorithms to match the corresponding phonemes (in International Phonetic Alphabet format) within the transcribed text with their respective time locations in the speech waveform. Finally, the frame positions corresponding to the start and end timestamps of each phoneme are obtained, allowing for localization of the phoneme boundaries.

\noindent\textbf{Details of Integration with Existing Methods}
We integrate pre-trained models of existing methods with phoneme-based localization techniques to construct different implementations of linguistic-based key motion capture. To elaborate, we generate a complete motion sequence using the pre-trained model of each existing method. Then, we extract key motions from the complete sequence based on the temporal position from phoneme-based localization. Subsequently, the CMC module is trained to extend the key motions from different methods into complete, continuous facial mesh sequences.
For fair comparisons, the multi-modal motion decoder and loss calculation of the CMC module remain consistent with our method.

\begin{figure}[t]
    \centering
    \includegraphics[width=\linewidth]{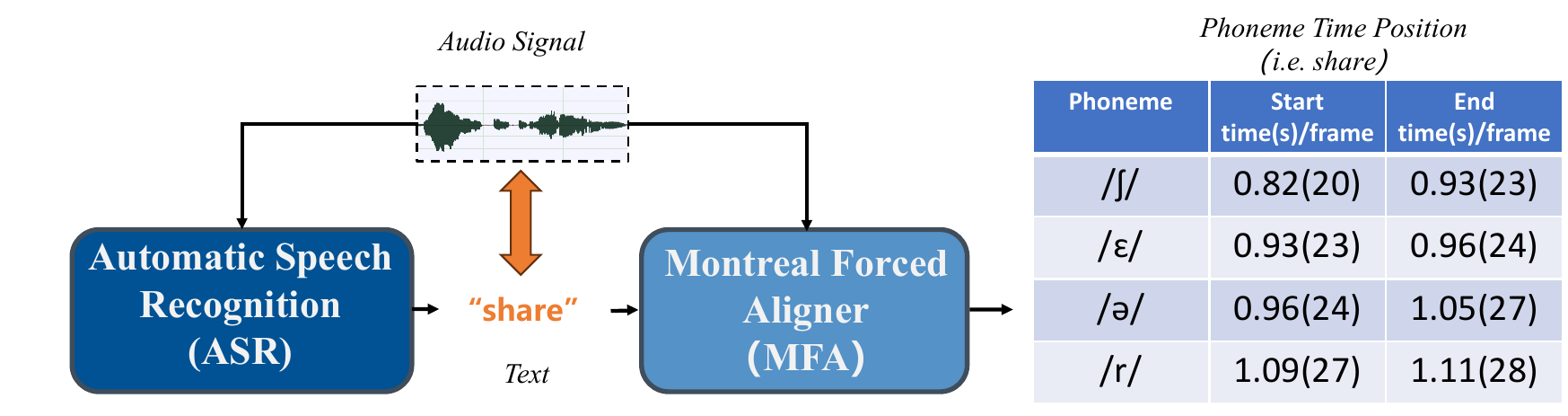}
    \caption{The pipeline of the Phoneme-based Localization Method includes the Automatic Speech Recognition (ASR) module and the Montreal Forced Aligner(MFA) module.
    }
    \label{fig:phoneme pipline}
    
\end{figure}

\section{Additional Results}
\label{SecAddResult}
\subsection{Visualization Results of Phoneme Boundaries}
\label{sec:phoneme visualization}
%
%
To better comprehend the prior that articulatory actions are more pronounced at phoneme boundaries and effectively capture the kinematic characteristics of the entire motion sequence, we visualized the pronunciation of words alongside their corresponding lip offsets.
We extracted audio fragments from the BIWI and VOCASET datasets, which are shown in Fig.~\ref{fig:visualization}.
We observed that the key motion positions determined by the Phoneme-based Localization Method approximately capture the inflection points of the lip movement curve, denoted as key points.
Once these key points are determined, the remaining frames can be effectively fitted using the linear interpolation method.
Therefore, these key points can well describe the patterns of lip movement.

\subsection{Integration with Existing Methods on VOCASET}
The results of integrating our proposed progressive learning mechanism utilizing key motion embeddings with existing methods on VOCA-Test are shown in Table \ref{tab:general_voca}.
The experimental results demonstrate that our proposed learning mechanism can achieve significant improvements over existing state-of-the-art methods\cite{fan2022faceformer, xing2023codetalker, peng2023selftalk} on VOCASET, further confirming the strong generalization capabilities of our design.
\begin{table}[t]
\centering
\caption{The results of integrating our proposed KMTalk with existing methods on VOCA-Test.}
\label{tab:general_voca}
\renewcommand\arraystretch{1}{%
\scalebox{1}{
\begin{tabular}{clcc}
\hline
\toprule
\multirow{2}{*}{Methods}                              &          & LVE$\downarrow$                              & FDD$\downarrow$                              \\
                                                      &          & \multicolumn{1}{l}{$\times 10^{-5}{\rm mm}$} & \multicolumn{1}{l}{$\times 10^{-7}{\rm mm}$} \\ \hline
\multirow{2}{*}{FaceFormer~\cite{fan2022faceformer}}  & Original & 4.1090                                       & 4.6675                                       \\
                                                      & After    & \textbf{3.9608}                              & \textbf{4.5343}                              \\ \hline
\multirow{2}{*}{CodeTalker~\cite{xing2023codetalker}} & Original & 3.9445                                       & 4.5422                                       \\
                                                      & After    & \textbf{3.8473}                              & \textbf{3.9043}                              \\ \hline
\multirow{2}{*}{SelfTalk~\cite{peng2023selftalk}}     & Original & 3.2238                                       & 4.0912                                       \\
                                                      & After    & \textbf{2.6608}                              & \textbf{3.6795}                              \\ \bottomrule \hline
\end{tabular}%
}
}
\end{table}
\begin{table}[t]
\caption{Ablation study for our components on VOCA-Test. 
}
\label{tab:ablation study voca}
\renewcommand\arraystretch{1.2}{%
\scalebox{0.8}{
\begin{tabular}{cccccc}
\hline
\toprule
\textbf{Phoneme-based}       & {\textbf{Key Motion-focused}} & \textbf{Audio Guidance} & \textbf{} & \multirow{2}{*}{\textbf{\textbf{LVE$\downarrow$}}} & \multirow{2}{*}{\textbf{\textbf{FDD$\downarrow$}}} \\ 
\textbf{Localization Method} & \textbf{Decoder}                                  & \textbf{in CMC}          &           &                                                    &                                                    \\ \hline
---                   & ---                      & ---                 &           & 3.2238                                             & 4.0912                                            \\
                                             
---                 & \Checkmark                        & \Checkmark                 &           & 3.0987                                             & 4.1578                                             \\
\Checkmark                   & ---                      & \Checkmark                 &           & 2.8402                                             & 4.0482                                             \\
\Checkmark                   & \Checkmark                        & ---               &           & 4.7366                                             & 5.2046                                             \\ \Checkmark                   & \Checkmark                        & \Checkmark                 &           & \textbf{2.2639}                                             & \textbf{4.0594} \\ \bottomrule \hline 
\end{tabular}%
}
}
\end{table}
\begin{table}[t]
\centering
\caption{Additional ablations on BIWI-Test-A dataset.}
\label{tab:ablation}
\renewcommand\arraystretch{1}
    {
    \scalebox{1}{
    \begin{tabular}{ccc}
    \hline
    \toprule
    \multirow{2}{*}{Ablation}      & LVE$\downarrow$          & FDD$\downarrow$          \\
                                 & $\times 10^{-5}{\rm mm}$ & $\times 10^{-7}{\rm mm}$ \\ \hline
    LKMA                         & \multirow{2}{*}{4.0604}  & \multirow{2}{*}{\textbf{2.3137}}  \\
    (wo/text loss and latent loss)   &                          &                          \\
    CMC                          & \multirow{2}{*}{4.1824}  & \multirow{2}{*}{3.3777}  \\
    (fusion with self-attention) &                          &                          \\ \hline
    \textbf{KMTalk(Ours)}        & \textbf{3.9654}          & 2.5446          \\ \bottomrule\hline
    \end{tabular}
    }}
\end{table}
\subsection{Additional Results}
\label{sec:additional results}
\noindent\textbf{Ablation Studies on VOCASET}
Ablation studies of KMTalk on VOCASET are presented in Table \ref{tab:ablation study voca}, and the results are consistent with the experiments conducted on BIWI. 
This further validates the effectiveness of the Phoneme-based Localization Method, Key Motion-focused Decoder, and Audio Guidance in CMC.

\noindent\textbf{Ablation Studies of Loss Functions}
The latent consistency loss, measured by MSE, aligns latent audio features with lip encoder outputs, enhancing feature consistency. The text consistency loss, quantified by CTC, ensures lip movements match the source audio for accurate lip-reading. We empirically found that with the current weight strategy, the re-weighted losses are comparable, achieving the optimizal results. Ablation studies in Table~\ref{tab:ablation} indicated a decrease in LVE without the use of these two losses, underscoring the importance of text and latent consistency loss for lip-reading accuracy. 

\noindent\textbf{Effectiveness of Fusion Module Design}
To validate the design of the CMC module, we conducted an ablation study that directly utilized a self-attention for multimodal fusion. The experimental results, presented in the third row of Table~\ref{tab:ablation}, indicate a decline in performance when using self-attention for multimodal fusion.
\subsection{Results of Key Motions Quantity}
\label{sec:key motion quantity}
The comparison results of key motion quantity are shown in Table \ref{tab:Comparison Results}. 
The results indicate that the quantity of key motions obtained with a uniform sampling stride of 3 is closest to the quantity obtained with the Phoneme-based Localization Method. 
Additionally, the experimental results suggest that uniform sampling does not consider the varying importance of different elements, and simply increasing or decreasing the quantity of key motions does not significantly improve the results. 
Therefore, proposing a prior to capture the varying importance of different elements is crucial for enhancing the model's performance.
\begin{figure*}[t]
    \centering
    \includegraphics[width=\linewidth]{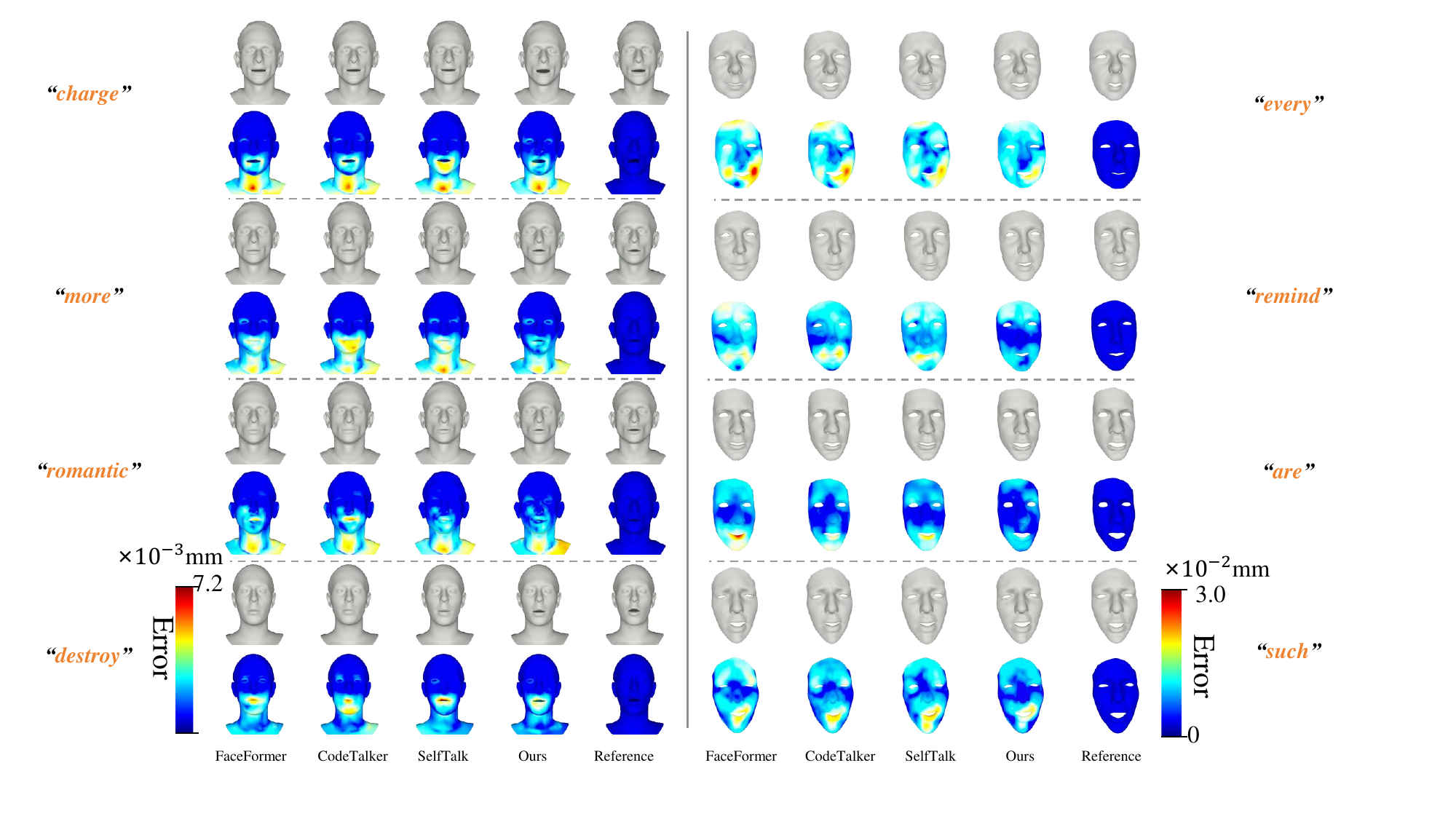}
    \caption{Qualitative comparisons on VOCASET (left) and BIWI (right).
    We provide visual comparisons of facial animations synchronized with eight syllables extracted from the test speech sequences.
    The 1st, 3rd, 5th, and 7th rows display synthesized meshes and their corresponding ground-truths, while the 2nd, 4th, 6th, and 8th rows visualize the L2 loss for individual frames.
    Our method demonstrates more precise mouth movement and generates more natural and synchronized motion sequences visually.
    }
    \label{fig:comparsion}
\end{figure*}
\begin{table}[t]
\caption{Comparison Results of Key Motions Quantity on BIWI-Test-A.}
\label{tab:Comparison Results}
\renewcommand\arraystretch{1.2}{%
\scalebox{0.8}{
\begin{tabular}{ccccc}
\hline
\toprule
\multirow{2}{*}{Method}   & \multirow{2}{*}{Quantity} & \multirow{2}{*}{Proportion} & LVE $\downarrow$                 & FDD $\downarrow$                 \\
                          &                           &                             & $\times 10^{-4}{\rm mm}$         & $\times 10^{-5}{\rm mm}$         \\ \hline
Uniform Sampling (Step 2) & 1944                      & 50.1\%                      & 4.1605                           & 3.0792                           \\
Uniform Sampling (Step 3) & 1301                      & 33.5\%                      & 4.1648                           & 2.8713                           \\
Uniform Sampling (Step 4) & 980                       & 25.3\%                      & 4.1655                           & 2.7521                           \\
Phoneme-based             & \multirow{2}{*}{1262}     & \multirow{2}{*}{32.5\%}     & \multirow{2}{*}{\textbf{3.9742}} & \multirow{2}{*}{\textbf{2.5973}} \\
Localization Method       &                           &                             &                                  &                                  \\ \bottomrule \hline
\end{tabular}%
}
}
\end{table}
\subsection{Additional Quantitative Comparisons}
\label{sec:comparsion}
Additional visual comparisons of facial meshes generated by various methods and ground truths are presented in Fig.~\ref{fig:comparsion}. Our method consistently shows lower errors across diverse speech sequences, underscoring its proficiency in producing more accurate facial animations.
\subsection{Visualization of Long Sequence Generation}
\begin{figure}[t]
    \centering
    \includegraphics[width=\linewidth]{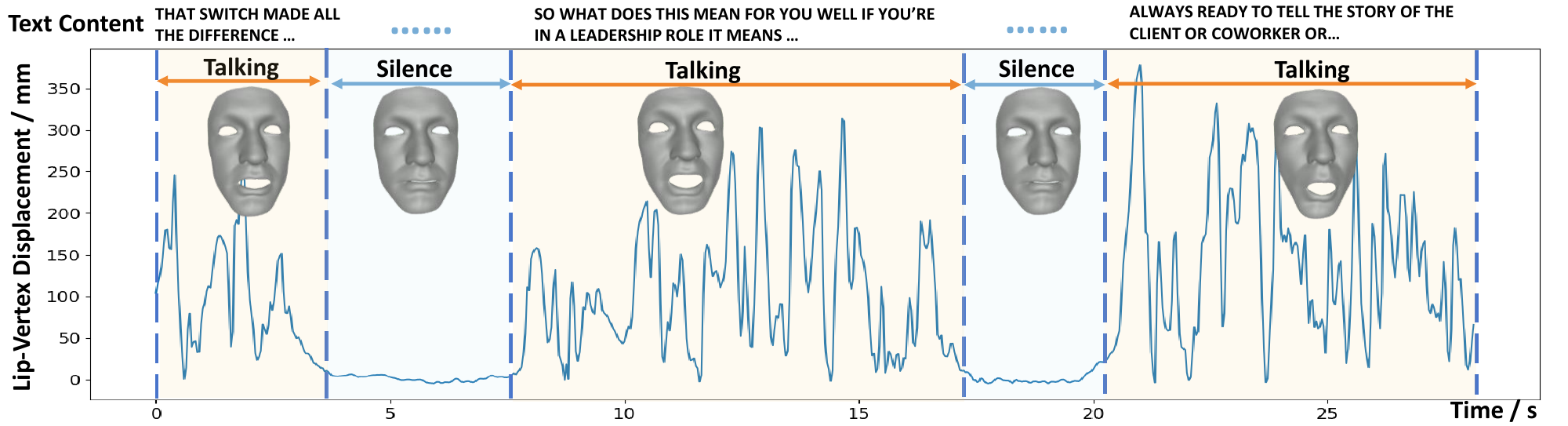}
    \caption{Test the longer audio clips with pauses}
    \label{fig:long}
\end{figure}

\begin{figure*}[h]
    \centering
        \subfloat[Visualization results on BIWI.]{\label{fig:sub1}
        \includegraphics[width = 1\linewidth]{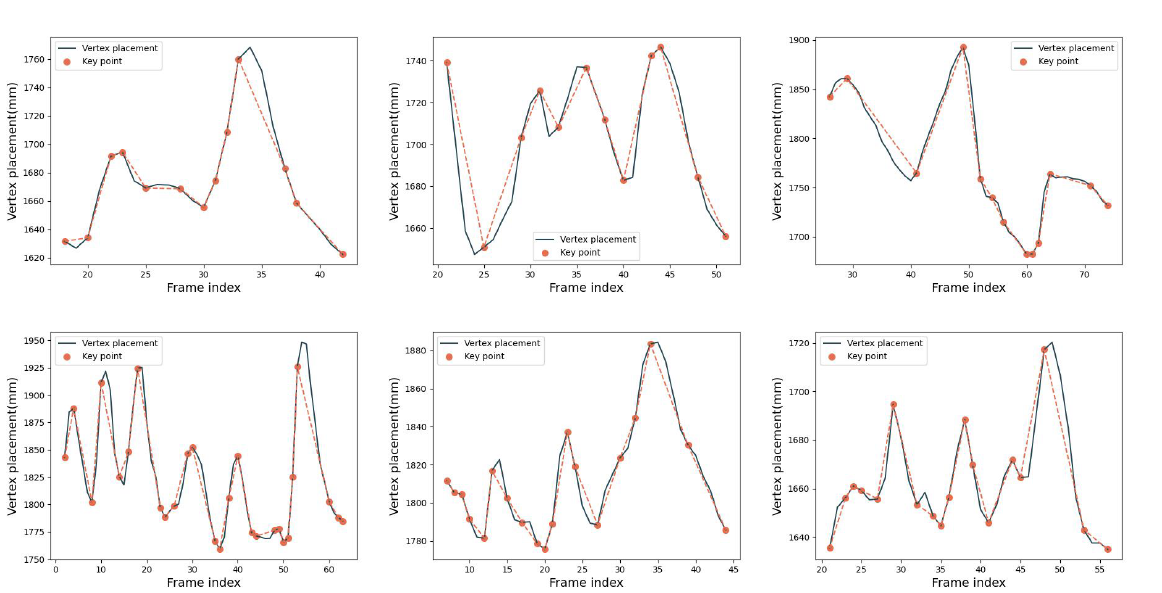}}
        \hfill
        \subfloat[Visualization results on VOCASET.]{\label{fig:sub2}
        \includegraphics[width = 1\linewidth]{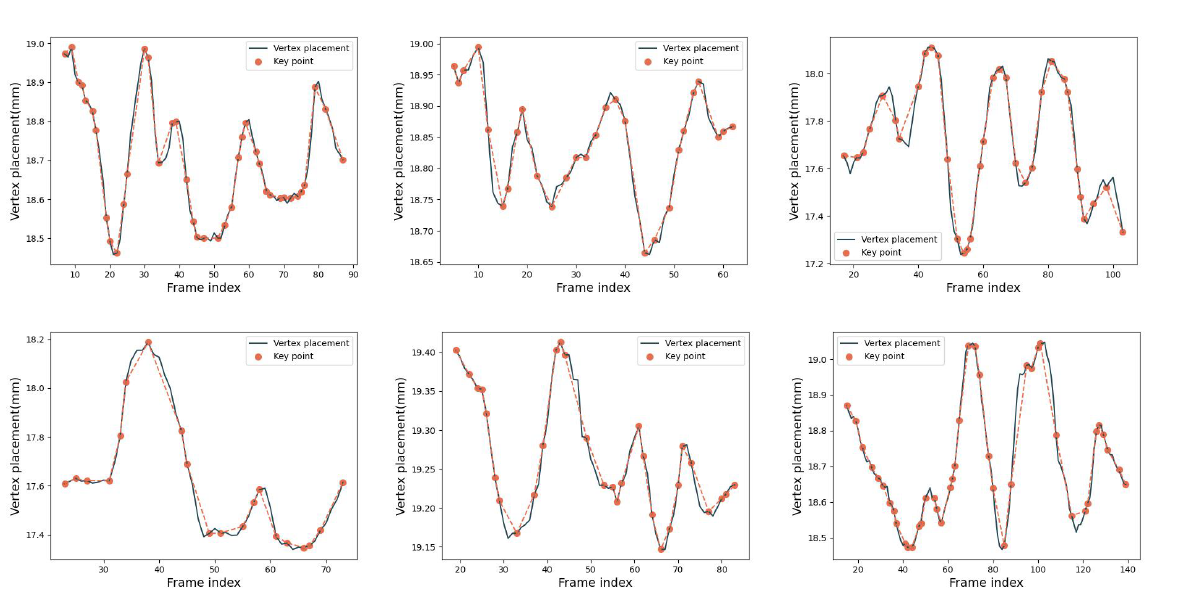}}
        \hfill

    \hfil
    \caption{The visualization of phoneme boundaries on the BIWI and VOCASET datasets is presented separately in (a) and (b).  Specifically, in this visualization, the vertex placement represents the cumulative Euclidean distance between the facial animation and the template in the lip region for each frame. The positions of key points are determined by the Phoneme Localization Method. Once these key points are marked, a linear interpolation method is employed to fit an approximate curve that closely approximates the marked key points.
    }
    \label{fig:visualization}
\end{figure*}

Both the VOCASET and BIWI datasets feature single-sentence inputs, typically under 5 seconds.  Although we follow prior works~\cite{peng2023selftalk,xing2023codetalker,fan2022faceformer} in experimenting with sentence-level datasets, our method can naturally extend to long sequences. We evaluated audio sequences including pauses with a duration of 1.5 minutes, and visualized the initial 30 seconds of intermediate frames and lip vertex displacement in Fig.~\ref{fig:long}. Our results can still produce accurate 3D talking face animation.For much longer audios, we segmented sequences into several clips and performed model inference for each clip individually.
\subsection{User Study}
\label{sec:user study}
\begin{figure*}[h]
    \centering
    \includegraphics[width=\linewidth]{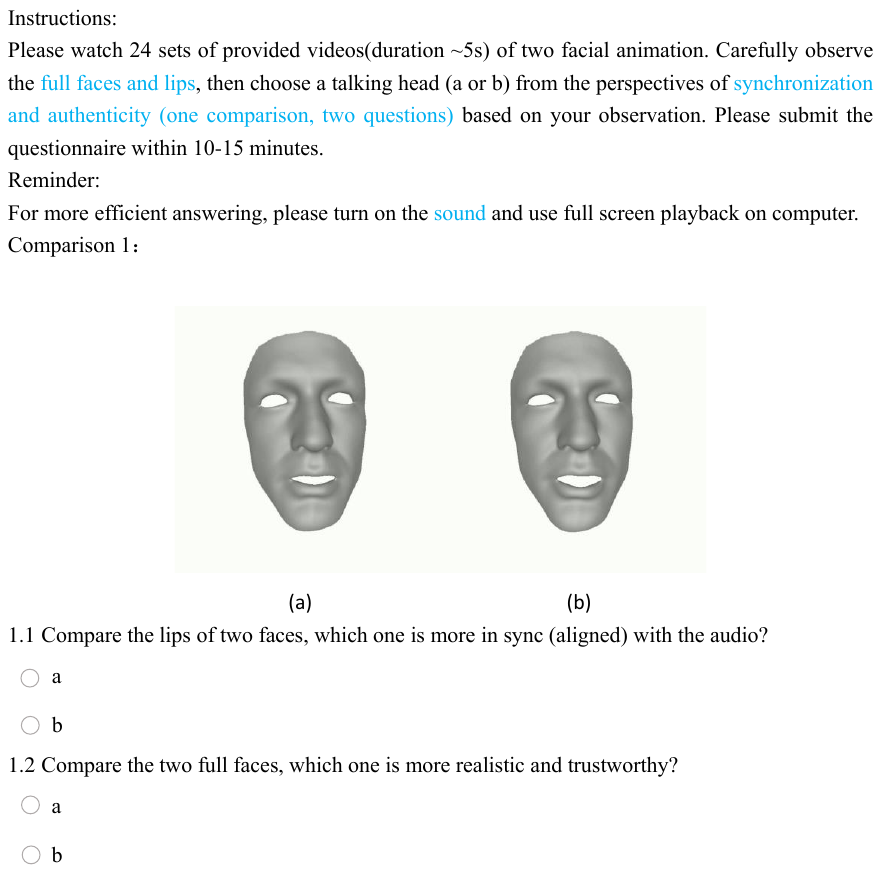}
    \caption{Designed user study interface. Each participant need to answer 24 video pairs and here only one video pair is shown due to the page limit.
    }
    \label{fig:userstudy}
\end{figure*}

The user study interface, designed for this research, is depicted in Fig.~\ref{fig:userstudy}. The anticipated completion time for the user study is estimated to be between 10 to 15 minutes, considering 24 pairs of videos, each lasting 5 seconds, and three repetitions of watching. To mitigate the influence of random selection, we exclude comparison results completed in less than two minutes. Each participant is presented with the user study interface, which includes 24 video pairs. Participants are instructed to evaluate the videos twice, answering the following questions for each pair: "Compare the lips of the two faces: which one is more in sync (aligned) with the audio?" and "Compare the two full faces: which one is more realistic and trustworthy?". The user study interface facilitates the evaluation process, allowing participants to make informed judgments based on these specific criteria.
\subsection{Video Comparison}
\label{sec: video comparison}
To better evaluate the qualitative results produced by both our KMTalk and competing methods, we provide a supplementary video for demonstration and comparison. Specifically, we utilize a variety of audio clips to test our model, including segments extracted from TED videos, audio sequences from the VOCASET and BIWI datasets, as well as speech extracted from supplementary videos of previous methods. The video demonstrates the capability of KMTalk to synthesize facial animations with realistic and natural lip synchronization. It is worth noting that in comparison to competing methods such as FaceFormer \cite{fan2022faceformer}, CodeTalker \cite{xing2023codetalker}, and SelfTalk \cite{peng2023selftalk}, which have experienced issues with over-smoothing, our KMTalk generates more dynamic and realistic facial movements with better lip synchronization. Furthermore, we demonstrate facial animations for speaking in different languages, such as Spanish, German, French, and more. The supplementary video serves as a visual demonstration, enabling a comprehensive comparison of the capabilities and strengths of our KMTalk approach. It highlights the ability of KMTalk to generate high-quality facial animations that exhibit natural lip movements, providing a more convincing and immersive user experience.

\section{Additional Discussions}
\label{sec:discussion}
\subsection{Inference Time}
KMTalk's inference time on a single 3090 GPU for ASR~\cite{radford2023robust} is 0.07 seconds and for MFA is 0.2 seconds on the BIWI dataset, with the LKMA and CMC modules together taking 0.37 seconds. Therefore, the average inference time for one audio clip is approximately 0.64 seconds. In comparison, Selftalk's inference time is 0.2 seconds per audio clip. Despite this increase, it is relatively minimal considering the complementary benefits of the modules and the overall performance of the system.
\subsection{Frame Rate}
\label{sec:frame rate}
Our method, KMTalk, operates at a frame rate of 30 fps on VOCASET and 25 fps on BIWI, following the setting of previous methods~\cite{fan2022faceformer,xing2023codetalker,peng2023selftalk}. The Audio analysis from our experiments indicates that phonemes have an average duration of approximately 0.1 seconds, thereby a frame rate exceeding 10 fps is adequate for identifying phoneme positions. Existing datasets, such as BIWI and VOCASET, with frame rates greater than 25 fps provide sufficient resolution to distinguish different phonemes. While a high frame rate (e.g., 60 fps) increases the number of frames between keyframes, potentially affecting the model's performance, our designed CMC module introduces global audio information, effectively mitigating the adverse effects of sparser keyframes.
We compared S2L+S2D~\cite{nocentini2023learning} in our setting and also adapted KMTalk to operate at 60 fps, and the results, shown in Table~\ref{tab:framerate}, demonstrate the superiority of KMTalk over S2L+S2D~\cite{nocentini2023learning} on LVE and FDD metrics and confirm the robustness of our approach at higher frame rates.
\begin{table}[t]
\centering
\caption{Quantitative comparisons on VOCA-Test dataset.}
\label{tab:framerate}
\renewcommand\arraystretch{1}
    {
    \scalebox{1}{
    \begin{tabular}{cccc}
    \hline
    \multirow{2}{*}{\textbf{Method}} & \multirow{2}{*}{\textbf{FPS}} & \textbf{LVE$\downarrow$} & \textbf{FDD$\downarrow$} \\
                                     &                               & $\times 10^{-5}{\rm mm}$          & $\times 10^{-7}{\rm mm}$          \\ \hline
    S2L+S2D                          & 60                            & 3.6467       & 4.0738       \\
    KMTalk                           & 60                            & 2.3115       & 4.0669       \\ \hline
    \textbf{KMTalk}                  & 30                            & \textbf{2.2639}       & \textbf{4.0594}       \\ \hline
    \end{tabular}
    }}
\end{table}
\subsection{Limitation Discussion}
Our method requires the localization of keyframes, thus in challenging scenarios such as dialect variations, localization may involve standard keyframe detection errors. However, our method has demonstrated a certain degree of robustness even in the presence of deviation in keyframe localization. As shown in the last row of Table~\textcolor{red}{4} in Sec.~\textcolor{red}{4}, our method still outperforms Selftalk by 26\% in the FDD metric despite the presence of keyframe offset deviation. In the future, integrating advanced ASR technology could enhance the model's robustness and adaptability to various speech patterns.

\end{document}